\documentclass{ifacconf}

\usepackage{graphicx}      
\usepackage{natbib}        
\usepackage{todonotes}
\usepackage{amsfonts}
\usepackage{xcolor}
\usepackage{amsmath}
\usepackage[autostyle=true]{csquotes}

\definecolor{MyPurple}{HTML}{FFAEFF}
\definecolor{MyGreen}{HTML}{008F00}
\definecolor{MyCyan}{HTML}{006F6F}
\begin{document}
\begin{frontmatter}

\title{Estimating Semantic Ambiguity via Gaussian Context Distributions for VLM-Driven Traversability Analysis} 


\thanks[footnoteinfo]{This work has in part been funded by the Vinnova project ROBOBUILDER with reference number 2024-02476.\\
\copyright~2026 the authors. This work has been accepted to IFAC for publication under a Creative Commons Licence CC-BY-NC-ND.}


\author{Ramona Häuselmann}\textbf{,} 
\author{Mario A.V. Saucedo}\textbf{,} 
\author{Christoforos Kanellakis}\textbf{, and} 
\author{George Nikolakopoulos}

\address{Robotics \& AI Team, Department of Computer, Electrical and Space Engineering, Lule\r{a} University of Technology, Lule\r{a} SE-97187, Sweden. Corresponding author's e-mail: ramhau@ltu.se.}

\begin{abstract}                
Autonomous navigation in unstructured environments requires robust scene understanding, yet Vision-Language Models (VLMs) often suffer from semantic ambiguity, where conflicting predictions can lead to dangerous failures. 
To address this, we present a novel pipeline for vision-based traversability estimation that explicitly models contextual uncertainty. 
Our approach utilizes \enquote{Conceptual Anchoring} to ground open-vocabulary VLM predictions onto a continuous physical traversability scale. 
By formulating the model's responses as a Gaussian Context Distribution (GCD), we derive both a dense traversability map and a dense uncertainty map based on the statistical properties of the distribution. 
Experimental validation on the real-world GOOSE dataset demonstrates that our proposed uncertainty metric effectively correlates with sources of ambiguity, such as visual artifacts and mixed terrain overlap. 
The method exhibits competitive performance while offering the distinct advantage of providing statistical uncertainty estimates to address semantic ambiguity, enabling safer and more reliable autonomous behavior in complex outdoor settings.
\end{abstract}


\end{frontmatter}

\section{Introduction}
Autonomous robotic navigation in unstructured environments is essential across many robotics applications.
Traditional geometry-based approaches often struggle to interpret the rich visual and semantic complexity found in natural scenes, while semantic segmentation alone often falls short in assessing functional traversability, i.e., whether a region can actually be traversed by the robot.
From a human perspective, traversability intuitively relates to the navigational \enquote{affordances} of the terrain, inferred through clues in the environment that suggest what actions are possible (\cite{gibson2014theory}).
For example, humans instinctively understand that concrete offers a more stable surface for walking than loose gravel or mud.
This kind of semantic reasoning is essential when deploying robots in unstructured environments.
To bridge this gap, recent advancements in Vision-Language Models (VLMs) have made \enquote{common sense} scene understanding accessible to robotic systems.

Nevertheless, translating the knowledge of these models into actionable, dense traversability representations that can enable robotic agents to act with higher autonomy presents a series of challenges.
VLMs are known to be susceptible to domain  shifts (\cite{zhang_vision-language_2024}) and operate as \enquote{black boxes}, making it difficult to understand the nature of their errors.
While some models provide confidence scores to assist in uncertainty estimation, reliance on these raw metrics is problematic since they are an output of the model itself.
One key problem is the presence of semantic ambiguity.
For example, a model might assign high confidence scores to the same pixel location for two contradictory prompts (e.g., \enquote{grass} and \enquote{water}).
Traditional approaches often resolve this by choosing the highest score and treating it as a confident prediction.
This, however, ignores the underlying conflict.
On the other hand, a pixel with a single, moderate peak for one prompt may be statistically more reliable than a pixel with competing high peaks, yet traditional metrics might discard it as \enquote{low confidence}.

\begin{figure*}[ht]
    \centering
    \includegraphics[width=\linewidth]{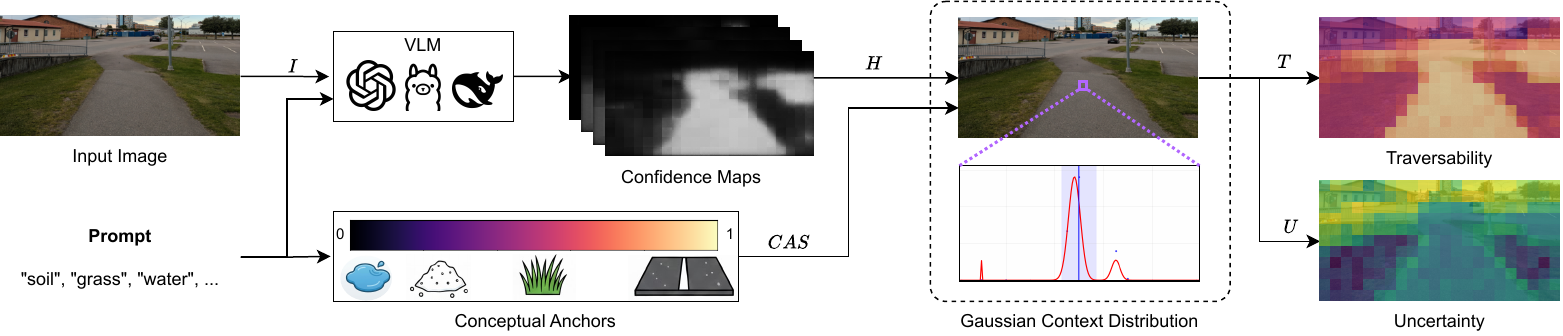}
    \caption{\textbf{Proposed method overview.} Given an input image a VLM segmentation model is used to infer per prompt confidence maps that are later fused with the proposed conceptual anchors in order to produce a per-patch Gaussian Context Distribution. From these distributions dense traversability and uncertainty maps are generated and used to determine the degree of visual ambiguity present in the environment.}
    \label{fig:pipeline}
\end{figure*}

This highlights a core limitation in current methods like DINO (\cite{liu_grounding_2024}) or SAM (\cite{li_semantic-sam_2023}), this is, the assumption that intra-prompt confidence (how confident the model is in a pixel relative to all other pixels sharing the same prompt) can be translated into inter-prompt confidence (the probability of a pixel belonging to a specific prompt versus other possible prompts).
By mixing these confidence metrics, standard approaches obscure important sources of uncertainty. 
This can lead to dangerous failures, where a hazardous region is misclassified as safe simply because its activation was marginally larger. 

In this work, we address this issue to enable more reliable, VLM-based dense traversability estimation.
To bridge the gap between intra- and inter-prompt confidence, we introduce a method that treats the model’s outputs as a distribution over traversability cues. 
Instead of forcing a hard choice between competing classes, our method explicitly captures multiple activation peaks. 
By evaluating this distribution statistically, we obtain a unified traversability estimate that reflects contributions from all relevant semantic contexts rather than a single dominant label.
This probabilistic view also provides a measure of uncertainty: the spread of the distribution serves as an indicator of ambiguity. 
Regions with high semantic ambiguity naturally yield higher uncertainty, allowing the robot to recognize and avoid areas where misclassification is likely.

\section{Related Work}
The robotics community has traditionally addressed terrain and traversability analysis from the perspective of feature extraction, three-dimensional mapping, and geometry (e.g. occupancy maps, slope, and height metrics). Recent works in traversability analysis for mobile robots have focused on: i) traversability frameworks using deep-learning networks~(\cite{e2025terrain}), ii) terrain classification approaches on sensor fusion (e.g. LiDAR-IMU)~(\cite{e2025terrain}) and iii) traversability estimation leveraging the recently developed VLMs~(\cite{weerakoon2025behav}). In~(\cite{saucedo_eat_2024}) the authors present a traversability method, based on a multi-modal fusion mechanism that combines geometric (surface normals) and semantic features (U-Net type segmentation). Moreover, the resulting terrain confidence was integrated with a control architecture for reactive navigation. In~(\cite{gasparino_wayfaster_2024}),
the authors present a self-supervised neural network for traversability prediction based on RGB and depth image fusion. The scheme was trained using data from a receding horizon estimator, eliminating the need for hand-engineered heuristics. In~(\cite{ruetz_foresttrav_2024}) the authors present ForestTrav, a LiDAR only traverability estimation leveraging 3D voxel representations. The fusion technique uses the LiDAR measurements to generate per-voxel statistics, combined with structural context and compactness of sparse convolutional neural networks (SCNNs). In~(\cite{adarsh2024}) the authors present a navigation architecture that uses Vision Language Models (VLMs) to identify the environmental context and to compute suitable, context-based trajectories. CoNVOI utilizes a multi-modal visual marking approach where obstacle-free regions in the RGB image are annotated with numbers, correlated with a local occupancy map, which grounds image locations and focuses the VLM's attention on navigable areas. In~(\cite{mohamed2025}) the authors propose VLM-GroNav, an autonomous navigation algorithm for outdoor environments based on VLM with physical grounding to handle diverse terrain traversability conditions. VLM-GroNav enhances the VLM's semantic understanding by fusing it with proprioceptive sensing, which provides direct, real-time measurements of intrinsic terrain properties like deformability and slipperiness using an in-context learning formulation. In~(\cite{daeun2025}) the authors present a multi-modal trajectory generation and selection algorithm for mapless outdoor navigation. The architecture combines the generation of candidate trajectories based on a Conditional Variational Autoencoder (CVAE) with a VLM-based trajectory selection, reasoning on human-like trajectory given the environmental context.






\textbf{Contributions:}
%
%
Based on the current state of the art, the key contributions and highlights of this work are as follows.
The main contribution of this paper is a pipeline for vision-based traversability estimation that explicitly addresses semantic ambiguity in Vision-Language Models by estimating contextual uncertainty. 
We introduce \textit{conceptual anchoring} to ground VLM assessments in physical material properties and model the responses as a \textit{Gaussian Context Distribution (GCD)}. 
This probabilistic approach allows us to derive dense traversability maps with fine-grained, per-patch scores while simultaneously extracting a contextual uncertainty map based on the statistical evaluation of the distribution's spread.
We demonstrate the functionality of our pipeline in unstructured real-world outdoor environments, providing a comprehensive evaluation that includes both qualitative analysis and quantitative comparisons against ground truth and state-of-the-art baselines.


\section{Problem Statement}
The objective of this work is to perform dense, vision-based and semantic ambiguity aware traversability analysis from a single image. 
Given an input RGB image $I \in \mathbb{R}^{H \times W \times 3}$, the problem is to estimate a corresponding dense traversability map $T$ and uncertainty map $U$.
Formally, we seek to find a mapping function $f$ such that: 
$$(T,U) = f(I)$$ 
where the output $T \in [0, 1]^{H \times W}$ and $U \in [0, 1]^{H \times W}$ are a map of the same spatial dimensions as the input. 
Each element $T_{i,j}$ in this map represents a quantitative, continuous traversability score for the pixel at spatial coordinates $(i, j)$. 
This score defines the ease and safety of traversal, where $T_{i,j} = 0$ signifies a completely untraversable region (e.g., water) and $T_{i,j} = 1$ signifies an optimally traversable region (e.g., asphalt).
Each element $U_{i,j}$ in this map represents a quantitative, continuous uncertainty value for the pixel at spatial coordinates $(i, j)$. 
This value measures the uncertainty of the traversability assessment, where $U_{i,j} = 0$ signifies a low uncertainty and $U_{i,j} = 1$ signifies a highly uncertain traversability estimation due to the presence of semantic ambiguity.


\section{Methodology}
\label{sec:methodology}
We present our pipeline for VLM-driven visual-based traversability analysis. 
The primary objective is to generate a dense, pixel-wise segmentation map where each pixel is assigned a traversability score and uncertainty estimation, directly enabling terrain-aware robot navigation.
We enhance a pre-trained, open-vocabulary segmentation model 
by providing it with explicit semantic traversability context.
Fig.~\ref{fig:pipeline} shows an overview of the proposed method.

\textbf{Traversability Context via Conceptual Anchoring:}
\label{sec:cas}
To provide the Vision-Language Model (VLM) with essential semantic context for traversability, we employ a \textit{Conceptual Anchor Set (CAS)}. 
This set is composed of descriptive text phrases that characterize various ground materials (e.g., \enquote{gravel}, \enquote{grass}, \enquote{sand}).

The core of this method is to explicitly ground the VLM's understanding of these concepts onto a defined traversability scale.
The CAS is structured as an ordered collection of concepts, arranged in ascending order based on the safety and ease of traversal associated with each description.
This ordering is mapped to a normalized scale from [0, 1], where 0 signifies completely untraversable terrain (e.g., \enquote{water}) and 1 represents optimal, highly traversable ground (e.g., \enquote{asphalt}).
The method allows for the precise specification of where each concept $c_i$ is to be placed on this [0, 1] continuum. 
We refer to this specified value as the concept's \textit{conceptual anchor position} $p_i$.

We formally define this traversability context set $CAS$ as an ordered set of tuples:
\begin{equation*}
    CAS=\{(c_1,p_1), (c_2, p_2), \dots ,(c_n, p_n)\}
\end{equation*}
where $c_i$ is the $i$-th descriptive text concept and $p_i$ is its corresponding conceptual anchor position on the traversability scale, such that $0 \leq p_i \leq 1$.
The set is ordered such that $p_1 \leq p_2 \leq \dots \leq p_n$.
This anchored set provides a structured, explicit context for reasoning about terrain safety.

\textbf{Per-Patch Context Distribution Estimation:}
\label{sec:per_patch_context_distribution}
To derive a dense traversability map $T$, we first process the input image $I$ using the VLM segmentation model, conditioned on our Conceptual Anchor Set, $CAS = \{(c_k, p_k)\}_{k=1}^n$.
This process generates a stack of $n$ semantic similarity heatmaps, $H=\{H_1, \ldots, H_n\}$, where each heatmap 
$H_k \in \mathbb{R}^{H_W \times W_W}$ 
corresponds to one of the $n$ concepts $c_k$.
$H_W$ and $W_W$ are the spatial dimensions of the VLM output, which correspond to the patches of the original image.
Each value $H_{k, (i,j)}$ in a heatmap represents a raw matching score (logit) indicating how strongly the image patch at coordinate $(i, j)$ aligns with the semantic concept $c_k$.

For each individual patch $(i, j)$, we extract its vector of $n$ raw scores, $\mathbf{s}_{(i,j)} = [s_1, \ldots, s_n]^\top$, where $s_k = H_{k, (i,j)}$. 
To convert these scores into a coherent probabilistic weighting, we apply a softmax function. 
This yields a normalized probability distribution $\mathbf{w}_{(i,j)}$ across the $n$ concepts:
\begin{equation*}
    w_{k, (i,j)} = \frac{e^{s_k}}{\sum_{m=1}^n e^{s_m}}
\end{equation*}
where $w_{k, (i,j)}$ is the resulting probability for concept $c_k$ at patch $(i, j)$.
Each weight $w_{k, (i,j)}$ is now directly associated with its corresponding conceptual anchor position $p_k$ from the $CAS$.
This set of weighted anchor positions, $\{(w_k, p_k)\}_{k=1}^n$, defines a discrete probability distribution for the traversability of patch $(i, j)$, which we introduce as the \textit{Gaussian Context Distribution (GCD)}.
We formulate the GCD as a Gaussian Mixture Model (GMM) defined over the traversability scale $x \in [0,1]$.
In this formulation, each concept $c_k$ acts as a component Gaussian centered at its anchor position $p_k$, weighted by the VLM's confidence $w_{k, (i,j)}$. 
We interpret the global mean of the GCD as the traversability score, and the global standard deviation as the uncertainty estimate.
\begin{itemize}
    \item GCD Mean ($\mu_{(i,j)}$): This is the traversability score. It represents the expected value of the mixture, integrating semantic evidence from all anchors.
    \item GCD Standard Deviation ($\sigma_{(i,j)}$): This quantifies the context uncertainty. A high $\sigma_{(i,j)}$ indicates that the GCD is multimodal or widely spread, implying the VLM is detecting conflicting concepts (e.g., evidence for both \enquote{mud} and \enquote{gravel} at different ends of the scale), while a low $\sigma_{(i,j)}$ indicates a confident consensus on the terrain type.
\end{itemize}
By calculating these moments for all patches, we generate two final dense maps: a traversability score map $T$, where $T_{i,j} = \mu_{(i,j)}$, and a context uncertainty map $U$, where $U_{i,j} = \sigma_{(i,j)}$.

\begin{figure*}[ht]
    \centering
    \includegraphics[width=0.75\linewidth]{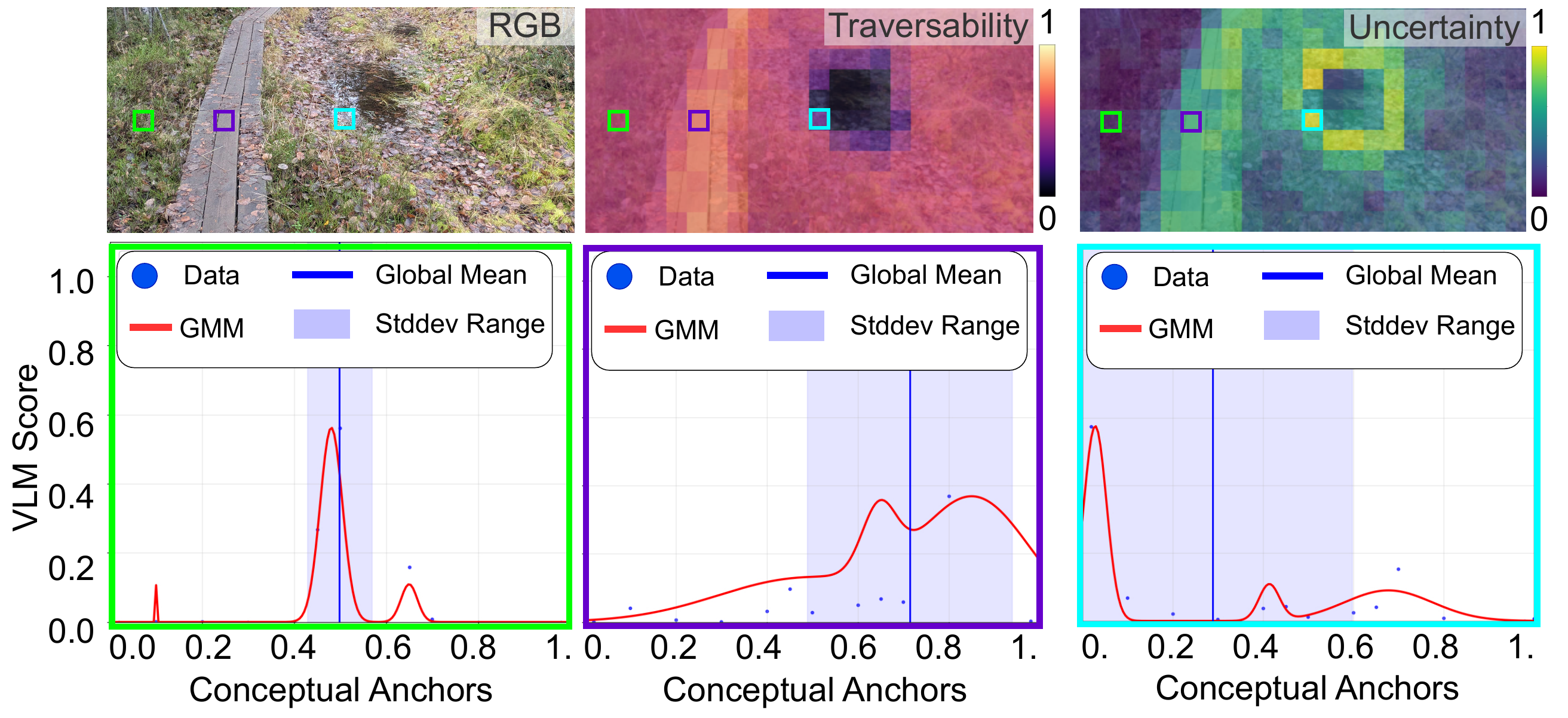}
    \caption{Top Row: An RGB input image followed by the predicted traversability and uncertainty maps. Three specific locations are highlighted: grass (green), wooden plank (purple), and puddle border (cyan). Bottom Row: The corresponding context distributions for the three highlighted locations. 
    }
    \label{fig:ex_single}
\end{figure*}

\section{Experimental Evaluation}
We validate our method on the real-world GOOSE dataset (\cite{goose-dataset}), utilizing CLIPSeg (\cite{luddecke_image_2021}) as our VLM backbone due to its open-vocabulary segmentation capability.
For our experiments we consider the following set of conceptual anchors 
\textit{CAS=\{(\enquote{water}, 0.02), (\enquote{mud}, 0.1), (\enquote{ice}, 0.2), (\enquote{snow}, 0.3), (\enquote{sand}, 0.4), (\enquote{moss}, 0.45), (\enquote{grass}, 0.5), (\enquote{gravel}, 0.6), (\enquote{soil}, 0.65), (\enquote{dirt}, 0.7), (\enquote{cobblestone}, 0.8), (\enquote{asphalt}, 0.98), (\enquote{concrete}, 1.0)\}}.
The experimental evaluation consists of a qualitative assessment of the predicted context distributions followed by a quantitative benchmarking of our traversability scores and uncertainty estimates against labeled ground truth and baselines as well as a qualitative validation of semantic ambiguity.

\subsection{Qualitative Evaluation}
Fig.~\ref{fig:ex_single} presents a qualitative evaluation of our traversability estimation framework, where the top row displays the input RGB image alongside the resulting traversability and uncertainty maps, while the bottom row details the context distributions for three distinct image locations (marked in green, purple, and cyan).
The obtained results illustrate the direct correlation between semantic ambiguity in the visual input and the uncertainty (represented by the standard deviation) of the distribution, for example:
\textbf{Low Ambiguity (Green)}: The area corresponding to the grass texture yields VLM activations that are tightly clustered. 
This lack of semantic ambiguity results in a unimodal distribution with a small standard deviation, indicating high confidence in the traversability estimate.\\
\textbf{Medium Ambiguity (Purple)}: The wooden plank area exhibits a wider range of VLM activations. This increased semantic variance leads to a broader distribution and, consequently, a higher standard deviation compared to the homogeneous grass area.\\
\textbf{High Ambiguity (Cyan)}: The border of the puddle presents the most complex scenario. Here, the presence of reflections and mixed textures triggers multiple, contradicting VLM activations. 
This is reflected in the distribution as a highly dispersed, multi-modal shape with a high standard deviation, capturing the high uncertainty of this traversability boundary.

\begin{figure}[hb]
    \centering
    \includegraphics[width=0.8\linewidth]{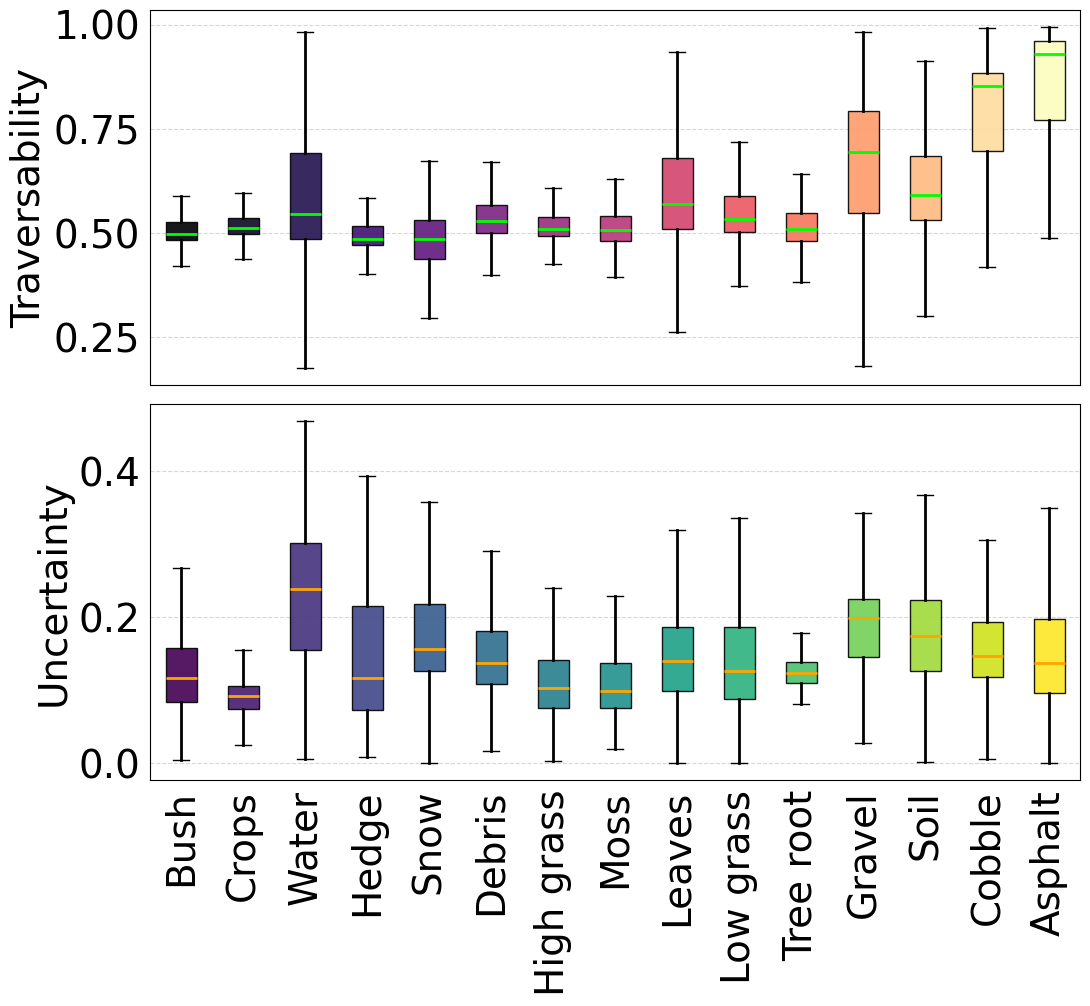}
    \caption{Boxplots of traversability and uncertainty on the GOOSE dataset, showcasing that prediction uncertainty correlates with semantic ambiguity.}
    \label{fig:ex_boxplots}
\end{figure}

\begin{figure*}[ht]
    \centering
    \includegraphics[width=0.8\linewidth]{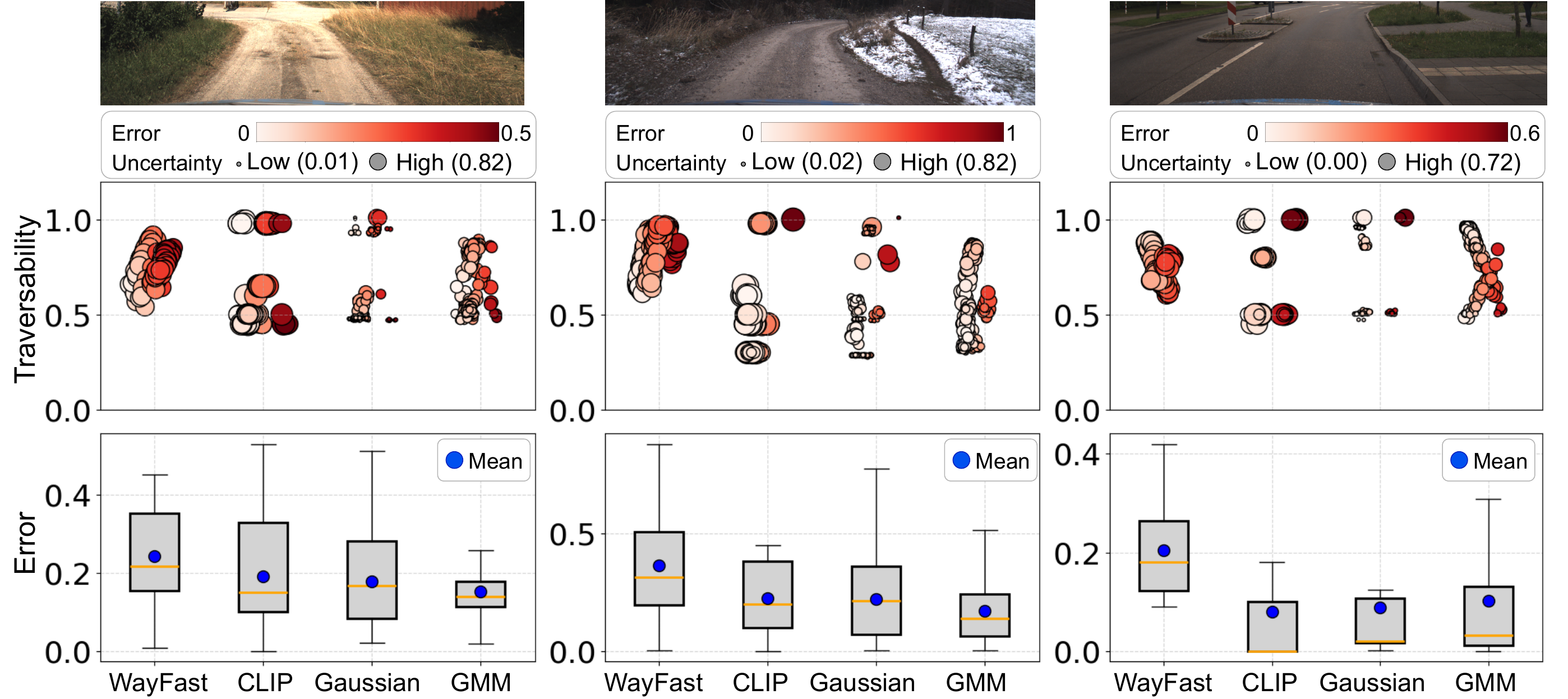}
    \caption{
    Comparative evaluation on the GOOSE validation split. 
    Top: RGB input scenarios. Middle: Predicted traversability swarm plots; horizontal jitter indicates error magnitude (lower error to left). Color: absolute error relative to the label-derived ground truth. Size: uncertainty (GMM/Gaussian: distribution spread; CLIP: inverse activation; WayFast: fixed). Bottom: Boxplots of error distributions by method.
    }
    \label{fig:ex_ablation}
\end{figure*}

\subsection{Groundtruth Comparison}
To assess the performance of the proposed method, we utilized the GOOSE dataset, which provides sensor data and ground truth annotations across a wide variety of outdoor environments.
Fig.~\ref{fig:ex_boxplots} illustrates the traversability scores (top) and their associated uncertainty (bottom) obtained with the proposed method for all relevant ground-truth terrain labels provided on the dataset.
We can observe how semantically ambiguous classes like \textit{water} exhibit a broader traversability distribution, which directly correlates to the degree of uncertainty, whereas more consistent terrains like \textit{grass} yield narrow distributions with low uncertainty. 
Aligning with human common sense, the traversability scores rank stable man-made surfaces like asphalt and cobble significantly higher (${T}>0.8$) than obstacles such as bush or hedge (${T}\approx0.5$). 
While also capturing the inherent unpredictability of unstructured terrains, where gravel and soil are predicted as highly traversable, but exhibit a higher uncertainty than asphalt, reflecting the \enquote{safe but uncertain} nature of unpaved surfaces. 
Finally, the significant upper whisker in the asphalt uncertainty distribution indicates that the method remains sensitive to visual anomalies rather than blindly assigning low uncertainty based on the class label alone.

\subsection{Baseline Comparison}
To compare our method against state-of-the-art baseline approaches, we utilized the validation split of the GOOSE dataset. 
As the dataset lacks native traversability scores, we established a quantitative ground truth by mapping the provided semantic segmentation labels to fixed traversability values. 
We compare our proposed GMM approach against three baselines: WayFast, a dedicated traversability framework (used here as a reference, noting that its internal definition of traversability may differ from our label-based proxy), CLIP, which assigns traversability based on the highest conceptual anchor activation, and a Gaussian baseline, representing an ablation of our method where the distribution is restricted to a single-modal fit.
Fig.~\ref{fig:ex_ablation} visualizes the performance across three sample scenarios, where the middle row displays the distribution of predicted traversability scores clustered by method. 
To visualize the error distribution, points with lower errors are positioned to the left of their cluster, and color intensity represents the magnitude of error relative to the ground truth. 
Point size correlates with uncertainty: for GMM and Gaussian, it represents the estimated standard deviation; for CLIP, it is inversely related to the activation score; while WayFast is rendered with a fixed size as it provides no uncertainty metric. 
The proposed method has comparable performance to baseline methods, while offering significant improvements in challenging scenarios.
By capturing the multi-modal nature of the data rather than enforcing a single peak, it achieves lower mean error and tighter error distribution, while retaining the stability of distribution modeling and the flexibility to handle semantic ambiguity.
%





\subsection{Semantic Ambiguity Analysis}
\begin{figure}
    \includegraphics[width=\linewidth]{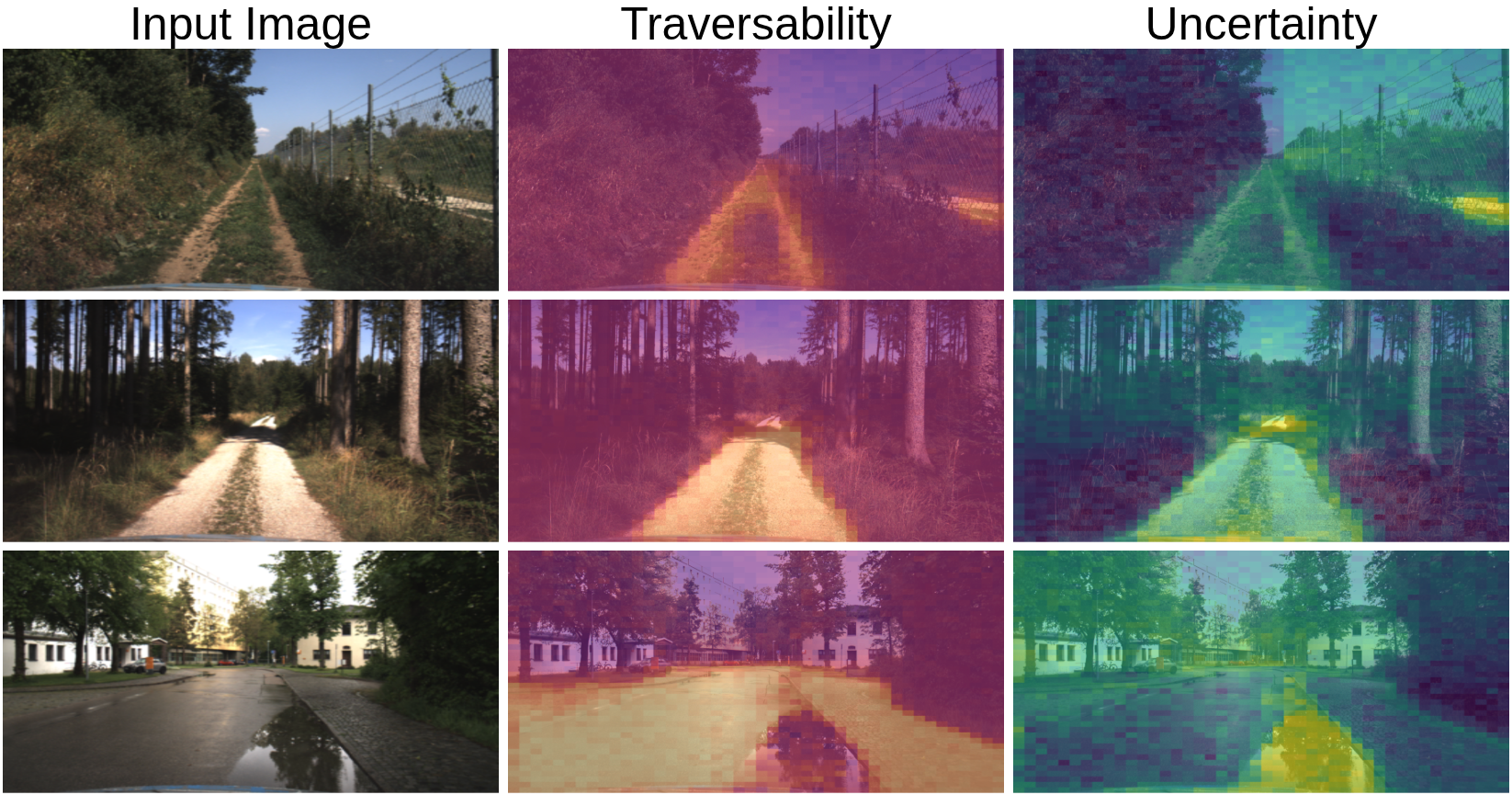}
    \caption{\textbf{Semantic ambiguity due to visual artifacts.} Our method's response to three distinct visual challenges: semi-transparent obstacles (top), variable illumination (middle), and surface reflections (bottom). 
    }
    \label{fig:visual_artifacts}
\end{figure}

During the experimental validation of the proposed method the following main sources of semantic ambiguity were identified:

\textbf{Visual Artifacts:} 
Fig.~\ref{fig:visual_artifacts} illustrates the impact of visual artifacts, including semi-transparent obstacles (fences), shadows, and reflections.
In the first example, although the terrain behind the fence is recognized as traversable by the proposed method, it is also classified as highly uncertain due to the presence of visual occlusion caused by the wired mesh. 
Likewise, in the second row, a shadowed path segment is classified as traversable, while presenting a high degree of uncertainty brought about by the lower visibility of the terrain. 
Finally, the last row showcases how environmental reflections on a puddle surface also produce high uncertainties values. 

\begin{figure}
    \includegraphics[width=\linewidth]{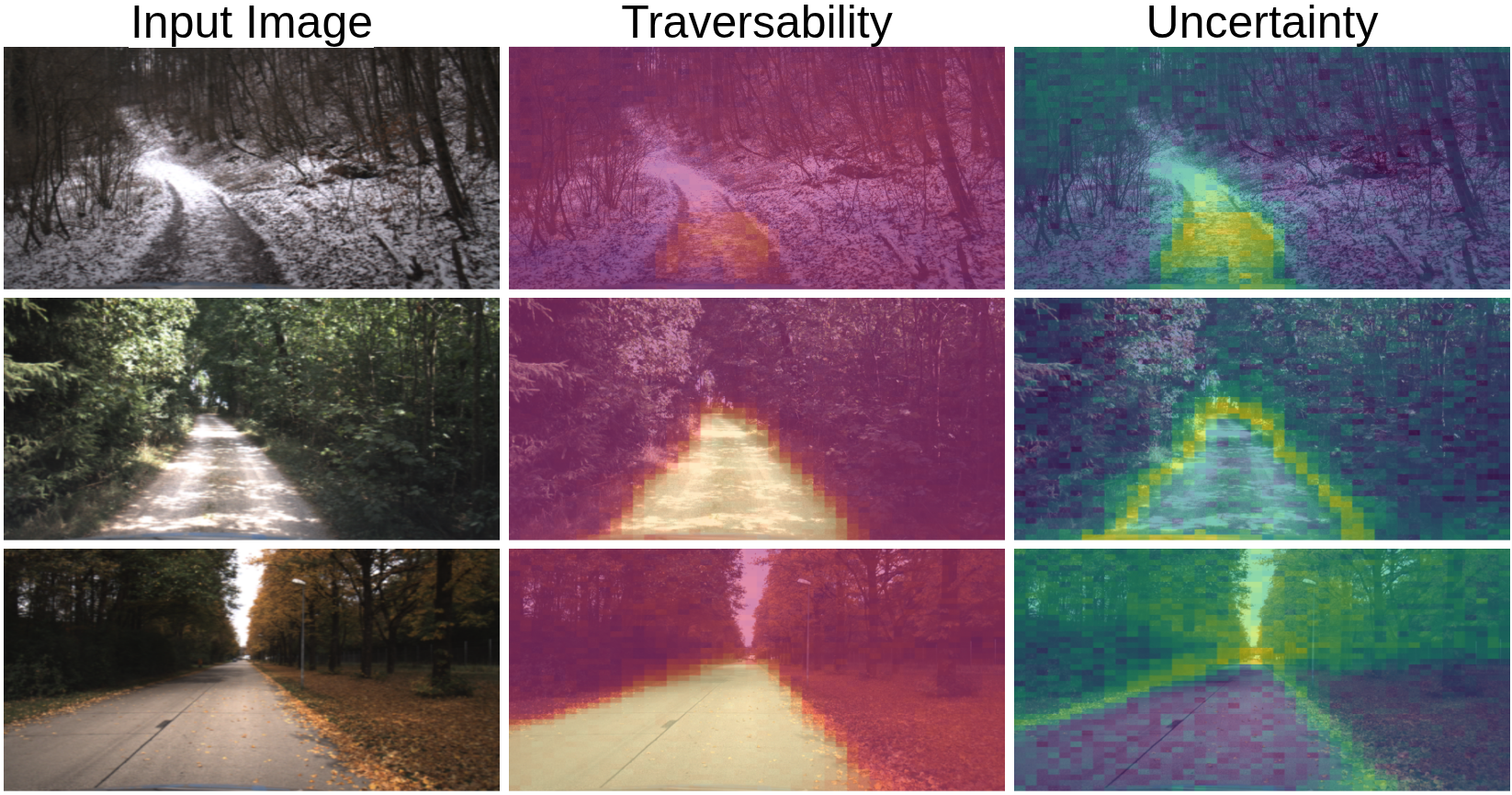}
    \caption{\textbf{Semantic ambiguity due to terrain overlap.} 
    Our method's response in scenes with partially overlapping terrains: light snow on gravel (top), encroaching vegetation (middle), and fallen leaves (bottom).
    %
    }
    \label{fig:terrain_overlap}
\end{figure}

\textbf{Overlapping terrain:}
Fig.~\ref{fig:terrain_overlap} showcases the method's performance in scenarios with overlapping of different terrains, such as snow on a gravel path, overgrowing vegetation at the edge of the road, and fallen leaves. 
The first row depicts how the exposed gravel is overall identified as traversable, while the zones where the snow covers the path exhibit a high uncertainty due to the terrain overlap. 
The second row highlights the challenge of overgrowing vegetation, where the system produces high uncertainty estimates along the road borders, where the vegetation obstructs the edge of the path. 
In the third row, the presence of fallen leaves on the road edge results in lower traversability estimates compared to the clean asphalt, alongside higher uncertainty values caused by the overlap between the two different terrains.

\subsection{Discussion}
The presented method draws from the assumption that semantic similarity in the VLM embedding space correlates with traversability.
However, this relationship does not strictly hold in all scenarios. 
A notable counter-example is the semantic proximity between \enquote{bush} and \enquote{grass}. 
While visually and semantically related in the CLIP embedding space, they represent opposing traversability states. 
This discrepancy can result in distributions with separated peaks and low-probability gaps, creating challenging modalities for the proposed method.
Despite these theoretical constraints, our experiments demonstrate that the method behaves robustly in diverse real-world cases. 
Nevertheless, scenarios where semantic and physical properties diverge suggest the need for complementary evaluation metrics to ensure reliability.

Future work could focus on refining uncertainty estimation through higher-order statistics (e.g., skewness and kurtosis) and exploring prompt-free traversability estimation directly in the feature space. 
Integrating this framework directly into a navigation stack presents another valuable research avenue, where uncertainty estimates could be utilized to enable safer, traversability-aware, and ambiguity-aware autonomous exploration.


\section{Conclusions}
In this work, we presented a novel pipeline for vision-based traversability estimation that explicitly addresses the semantic ambiguity inherent in Vision-Language Models. 
By introducing Conceptual Anchoring, we successfully grounded open-vocabulary VLM predictions into a continuous physical traversability scale. 
Furthermore, by modeling the VLM outputs as a Gaussian Context Distribution (GCD), we derived a robust mechanism to estimate both a dense traversability score and a corresponding pixel-wise uncertainty map.
Our experimental evaluation on the GOOSE dataset demonstrated that the proposed uncertainty metric effectively correlates with real-world sources of ambiguity, including visual artifacts like reflections, terrain overlaps, and semantic novelty. 
The method showed competitive performance against the baseline while offering the distinct advantage of statistically valid uncertainty estimation, which traditional maximization-based approaches fail to provide. 
Ultimately, this framework provides a crucial step toward safer autonomous navigation by enabling robots to distinguish between confirmed safe terrain and semantically ambiguous areas that require cautious behavior.

\section*{DECLARATION OF GENERATIVE AI AND AI-ASSISTED TECHNOLOGIES IN THE WRITING PROCESS}
During the preparation of this work the authors used Google Gemini in order to refine and grammar check the wording of this manuscript. After using this tool, the authors reviewed and edited the content as needed and take full responsibility for the content of the publication.

\bibliography{ifacconf}             
                                                   
\end{document}